\documentclass[10pt,twocolumn,letterpaper]{article}

\usepackage{iccv}
\usepackage{times}
\usepackage{epsfig}
\usepackage{graphicx}
\usepackage{amsmath}
\usepackage{amssymb}

\usepackage{dsfont}
\usepackage{algorithm}
\usepackage{algorithmic}
\usepackage{bm}
\usepackage{booktabs}
\usepackage{multirow}
\graphicspath{{figures/}}
\usepackage[breaklinks=true,bookmarks=false]{hyperref}

\iccvfinalcopy 

\def\iccvPaperID{1746} 

\ificcvfinal\fi

\begin{document}
	
\title{Bayesian Optimized 1-Bit CNNs}
	
\author{Jiaxin Gu,\textsuperscript{1,${\dag}$} Junhe Zhao,\textsuperscript{1,${\dag}$} Xiaolong Jiang,\textsuperscript{1,2,3} Baochang Zhang,\textsuperscript{1,*}\\
Jianzhuang Liu,\textsuperscript{4} Guodong Guo,\textsuperscript{2,3} Rongrong Ji\textsuperscript{5,6}\\
 \\
\textsuperscript{1}Beihang University, Beijing, China\\
\textsuperscript{2}Institute of Deep Learning, Baidu Research, Beijing China\\
\textsuperscript{3}National Engineering Laboratory for Deep Learning Technology and Application\\
\textsuperscript{4}Huawei Noah's Ark Lab, China\\
\textsuperscript{5}School of Information Science and Engineering, Xiamen University, Fujian, China\\
\textsuperscript{6}Peng Cheng Lab, Shenzhen, China\\
\textsuperscript{*}Corresponding author, email: bczhang@buaa.edu.cn\\
\textsuperscript{${\dag}$} Co-first author\\
}
	
\maketitle
\ificcvfinal\thispagestyle{empty}\fi
	
	\begin{abstract}
		Deep convolutional neural networks (DCNNs) have dominated the recent developments in computer vision through making various record-breaking models. However, it is still a great challenge to achieve powerful DCNNs in resource-limited environments, such as on embedded devices and smart phones. Researchers have realized that 1-bit CNNs can be one feasible solution to resolve the issue; however, they are baffled by the inferior performance compared to the full-precision DCNNs. In this paper, we propose a novel approach, called Bayesian optimized 1-bit CNNs (denoted as BONNs), taking the advantage of Bayesian learning, a well-established strategy for hard problems, to significantly improve the performance of extreme 1-bit CNNs. We incorporate the prior distributions of full-precision kernels and features into the Bayesian framework to construct 1-bit CNNs in an end-to-end manner, which have not been considered in any previous related methods. The Bayesian losses are achieved with a theoretical support to optimize the network simultaneously in both continuous and discrete spaces, aggregating different losses jointly to improve the model capacity. Extensive experiments on the ImageNet and CIFAR datasets show that BONNs achieve the best classification performance compared to  state-of-the-art 1-bit CNNs.
	\end{abstract}
	
	\begin{figure}[htb]
		\centering
		\includegraphics[width=1.0\linewidth]{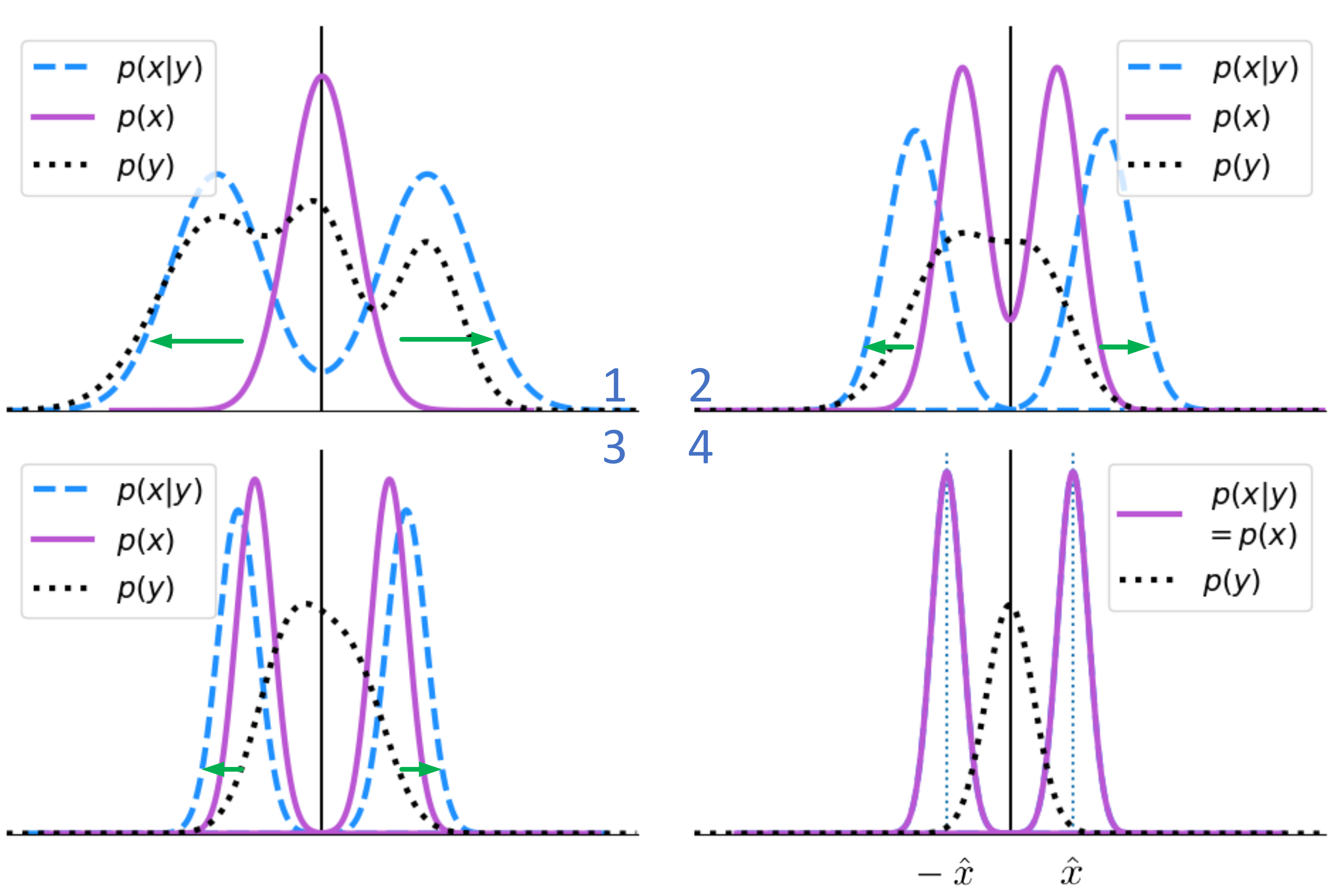}
		\caption{The evolution of the prior $p(\bm{x})$, the distribution of the observation $\bm{y}$, and the posterior $p(\bm{x}|\bm{y})$ during learning, where $\bm{x}$ is the latent variable representing the full-precision parameters and $\bm{y}$ is the quantization error. At the beginning, the parameters $\bm{x}$ are initialized according to a single-mode Gaussian distribution. When our learning algorithm converges, the ideal case is that (\romannumeral1) $p(\bm{y})$ becomes a Gaussian distribution $\mathcal N(0,\nu)$, which corresponds to the minimum reconstruction error, and (\romannumeral2) $p(\bm{x}|\bm{y})=p(\bm{x})$ is a Gaussian mixture distribution with two modes where the binarized values $\hat{\bm{x}}$ and $-\hat{\bm{x}}$ are located.}
		\label{figure:main-structure}
	\end{figure}
	
	\section{Introduction}
	
	Deep convolutional neural networks (DCNNs) have exhibited their superior feature representation power in both low-level \cite{dong2014learning,liu2018image} and high-level vision tasks \cite{he2016deep,liu2016ssd, wan2019c,wan2019min}. However, this superiority comes with prohibitive computation and storage overheads. In most cases, heavy parameters of DCNNs are stored as floating point numbers, each of which usually takes 32 bits, and the convolution operation is implemented as matrix multiplication between floating-point operands. These floating-point based operations are time-consuming and storage-demanding. Consequently, DCNNs are infeasible to be deployed on edge devices such as cellphones and drones, due to the conflict between high demands and limited resources. To tackle this problem, substantial approaches have been explored to compress DCNNs by pruning \cite{PruningFilter,han2015learning} or quantization \cite{paper14}.
	
	Quantization approximates full-precision values with lower-precision ones, therefore it can simultaneously accelerate the convolution operation and save storage expense. In particular, 1-bit convolution neural networks (1-bit CNNs) are the extreme cases of quantization, whose convolution kernels and activations are binarized, such as {$\pm1$} in \cite{paper10} or {$\pm \alpha_l$} in \cite{rastegari2016xnor}. Recently, DoReFa-Net \cite{zhou2016dorefa} exploits 1-bit convolution kernels with low bit-width parameters and gradients to accelerate both the training and inference phases. Differently, ABC-Net \cite{lin2017towards} adopts multiple binary weights and activations to approximate full-precision weights such that the prediction accuracy degradation can be alleviated. Beyond that, modulated convolutional networks are presented in \cite{cvprxiaodi2018} to only binarize the kernels, and achieve better results than the compared baselines. Leng \textit{et al}. \cite{leng2018extremely} borrows the idea from ADMM, which compresses deep models with network weights represented by only a small number of bits. Bi-real net \cite{liu2018bi} explores a new variant of residual structure to preserve the real activations before the sign function and proposes a tight approximation to the derivative of the non-differentiable sign function. Zhuang \textit{et al.} \cite{Zhuang_2018_CVPR} present $2\!\!\sim\!\!4$-bit quantization using a two-stage approach to alternately quantize the weights and activations, and provide an optimal tradeoff among memory, efficiency and performance. Furthermore, WAGE \cite{ICLR2018wu} is proposed to discretize both the training and inference processes, and it quantizes not only weights and activations, but also gradients and errors. In \cite{gu2018projection}, a quantization method is introduced based on a discrete back propagation algorithm via projection for a better 1-bit CNNs. Other practices are studied in \cite{tang2017train, alizadeh2018empirical,ding2019regularizing} with improvements over previous works.
	
	\begin{figure*}[htb]
		\centering
		\includegraphics[width=\linewidth]{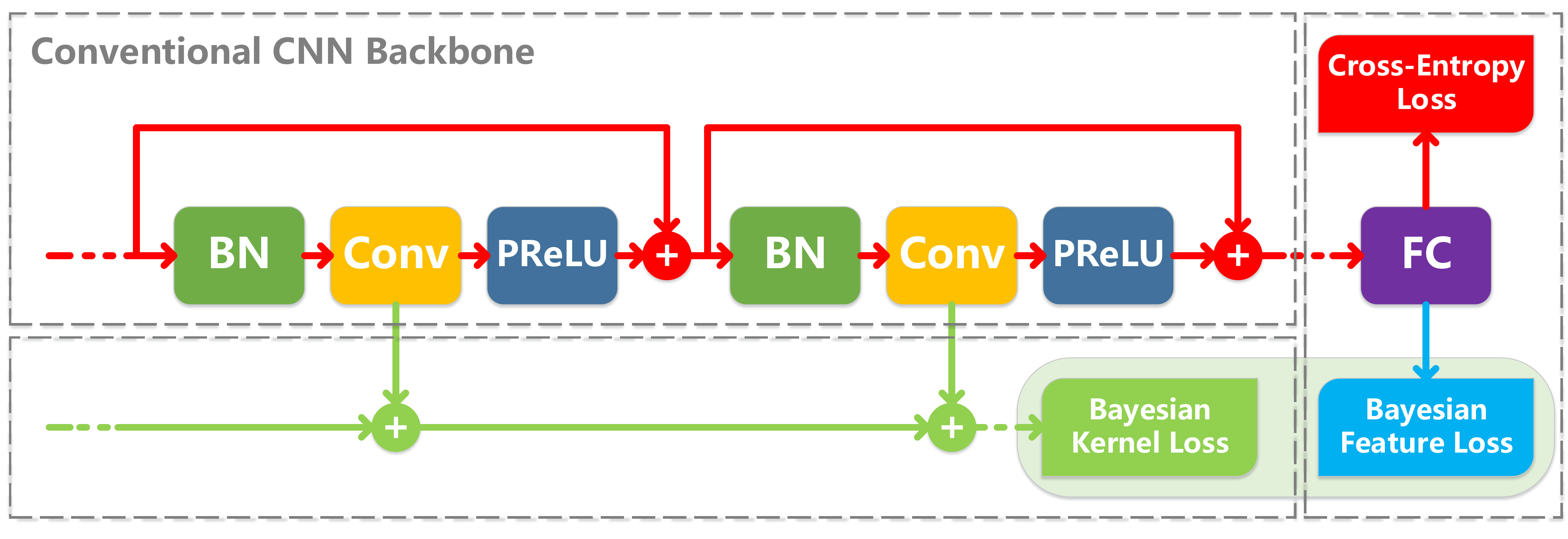}
		\caption{By considering the prior distributions of the kernels and features in the Bayesian framework, we achieve two new Bayesian losses to optimize the calculation of 1-bit CNNs. The Bayesian kernel loss improves the layer-wise kernel distribution of each convolution layer, while the Bayesian feature loss introduces the intra-class  compactness to alleviate the disturbance induced by the quantization process. Note that the Bayesian feature loss is only applied to the fully-connected layer.}
		\label{main-structure}
	\end{figure*}	
	
	Although these prevailing 1-bit CNNs use much less storage than conventional full-precision CNNs, yet compared to full-precision CNNs, they suffer from degraded accuracy in applications. Two reasons should account for this degradation: 1) the relationship between full-precision and 1-bit CNNs is not fully investigated for promoting the performance of 1-bit CNNs;  2) Bayesian learning, as a well-established strategy for global optimization \cite{mockus1978application,blundell2015weight}, is overlooked  in the field of 1-bit CNNs, although it can be  beneficial to the optimization of 1-bit CNNs according to our observations.
	
	In this paper, a Bayesian learning algorithm is proposed to optimize our 1-bit CNNs, leading to improved  accuracy and efficiency. Theoretically speaking, we achieve two novel Bayesian losses, with the help of Bayesian learning, to solve the difficult problem of CNNs binarization. {For 1-bit CNNs, the full-precision kernels  are  binarized to two quantization values (centers) gradually. Ideally, the quantization error is minimized when the full-precision kernels follow a Gaussian mixture model with each Gaussian centered at each quantization value. Given two  centers for 1-bit CNNs,  two Gaussians forming the mixture model are employed to model the full-precision kernels.}  The whole procedure can be illustrated by Fig.\;\ref{figure:main-structure}, when the learning algorithm converges with a binary quantization, the ideal result should be that: (1) the reconstruction error is minimized, and (2) the distribution of the parameters is a Gaussian mixture model with two modes centered at the binarized values separately. This assumption leads to our two new losses, referred to as the Bayesian kernel loss and Bayesian feature loss. The advantages of these novel losses are twofold. On one hand, they can be jointly applied with the conventional cross-entropy loss within the same back-propagation pipeline, such that the advantages of Bayesian learning is intrinsically inherited to optimize difficult problems. On the other hand, they can comprehensively supervise the training process of 1-bit CNNs with respect to both the kernel distribution and the feature distribution. 
	In summary, the contributions of this paper include:
	
	(1) We propose two novel Bayesian losses to optimize 1-bit CNNs, which are designed via exploiting Bayesian learning to fully investigate the intrinsic relationship between full-precision and 1-bit CNNs in terms of kernels and features.	
	
	(2) We develop a novel Bayesian learning algorithm to build 1-bit CNNs in an end-to-end manner. The proposed losses supervise the training process considering both the kernel distribution and the feature distribution, which are more comprehensive and efficient. 	
	
	(3) Our models achieve the best classification performance compared to other state-of-the-art 1-bit CNNs on the ImageNet and CIFAR datasets.
	
	\section{Proposed Method}
	Bayesian learning is one of the mainstreams in machine learning, which has been applied to building and analyzing neural networks to accomplish computer vision tasks \cite{blundell2015weight,mackay1992practical}.  In the paper, we leverage the efficacy of Bayesian learning to build 1-bit CNNs in an end-to-end manner. In particular, we lead to two novel Bayesian losses, based on which we optimize 1-bit CNNs with improved efficiency and stability. In a unified theoretical framework, these Bayesian losses not only take care of the kernel weight distribution specific in 1-bit CNNs, but also supervise the feature distribution. Fig.\;\ref{main-structure} shows how the losses interact with a CNN backbone. For clarity, in Table \ref{tab:variables} we first describe the main notation used in the following sections.
	
	\begin{table*}[htbp]
		\caption{A brief description of the main notation used in the paper.}
		\centering
		\begin{tabular}{l l l l}
			\toprule
			$\bm{X}^l_i$: full-precision kernel vector & $\bm{w}^l$: modulation vector & $\bm{\mu}_i^l$: mean of $\bm{X}_i^l$ & $\mathbf{\bm{\Psi}}^l$: covariance of $\bm{X}^l$  \\
			$\hat{\bm{X}}^l_{i}$: quantized kernel vector & $\bm{f}_m$: features of class $m$ & $\lambda$: trade-off scalar for $L_B$ & $\bm{c}_m$: mean of $\bm{f}_m$\\
			\hline
			$i$: kernel index & $l$: layer index & $m$: class index & $k$: dimension index\\
			$I_l$: number of kernels at layer $l$ &  $L$: number of layers & $M$: number of classes & $\nu$: variance of quantization error\\
			\bottomrule
		\end{tabular}
		\label{tab:variables}
	\end{table*}
	
	\subsection{Bayesian Losses}
	In state-of-the-art 1-bit CNNs \cite{leng2018extremely,rastegari2016xnor,gu2018projection}, the optimization involves in both continuous and discrete spaces. In particular, training a 1-bit CNN involves three steps: forward pass, backward pass, and parameter update through gradients. The binarized weights ($\hat{\bm{x}}$) are only considered during the forward pass (inference) and gradient calculation.  After updating the parameters, we have the full-precision weights ($\bm{x}$). As revealed in \cite{leng2018extremely,rastegari2016xnor,gu2018projection}, how to connect $\hat{\bm{x}}$ with $\bm{x}$ is the key to determine the network performance. In this paper, we propose to solve it in a probabilistic framework, in order to obtain optimal 1-bit CNNs.
	
	\textbf{Bayesian kernel loss.} We start with the fundamentals: given a parameter, we want it to be as close as possible before and after quantization, such that the quantization effect is minimized. Then, define
	\begin{equation}
	\bm{y} = \bm{w^{-1}}\circ\hat{\bm{x}} - \bm{x},
	\label{quantization of kernel}
	\end{equation}
	where $\bm{x},\hat{\bm{x}} \in \mathbf{R}^n$ are the full-precision and quantized vectors respectively, $\bm{w} \in \mathbf{R}^n$ denotes a learned vector to reconstruct $\bm{x}$, $\circ$ represents the Hadamard product, and $\bm{y}$ is the reconstruction error assumed to obey a Gaussian prior with zero mean and variance $\nu$. Given $\bm{y}$, we seek $\hat{\bm{x}}$ for binary quantization (1-bit CNNs) such that:
	\begin{equation}
	\hat{\bm{x}} =\max p(\bm{x}|\bm{y}),
	\label{bczhang}
	\end{equation}
	which indicates that under the most probable $\bm{y}$ (corresponding to $\bm{y}=\bm{0}$ and $\bm{x}=\bm{w}^{-1}\circ \hat{\bm{x}}$, {\it i.e.}, the minimum reconstruction error), the distribution of the latent variable $\bm{x}$ is a Gaussian mixture with two modes, locating at the quantization values, as shown in Fig.\;\ref{figure:main-structure}. And we have:
	\begin{equation}
	\begin{aligned}
	p(\bm{x|y}) &\propto \exp(-\frac{1}{2}(\bm{x}-\bm{\widetilde \mu})^T\bm{\Psi}^{-1}(\bm{x}-\bm{\widetilde \mu}))\\
	&+\exp(-\frac{1}{2}(\bm{x}+\bm{\widetilde \mu})^T\bm{\Psi} ^{-1}(\bf{x}+\bm{\widetilde \mu})),
	\end{aligned}
	\end{equation}
	where we set $\bm{\widetilde \mu} =  \bm{w^{-1}}\circ\hat{\bm{x}}$. However, Eq.\;\ref{bczhang} is difficult to solve. From a Bayesian perspective, we resolve this problem via maximum a posteriori (MAP) estimation:
	\begin{equation}
	\begin{aligned}
	\max p(\bm{x}|\bm{y}) &= \max p(\bm{y}|\bm{x})p(\bm{x})\\
	&=\min ||\hat{\bm{x}}-\
	\bm{w}\circ\bm{x}||^2_2 - 2\nu \log(p(\bm{x})), 
	\end{aligned}
	\label{type1}
	\end{equation}
	where
	\begin{equation}
	p(\bm{y}|\bm{x}) \propto \exp(-\frac{1}{2\nu}||\bm{y}||^2_2) \propto \exp(-\frac{1}{2\nu}||\hat{\bm{x}}-\bm{w}\circ\bm{x}||^2_2).
	\label{p(y|x)}
	\end{equation}
	In Eq.\;\ref{p(y|x)}, we assume that all the components of the quantization error $\bm{y}$ are i.i.d, thus resulting in such a simplified form. As shown in Fig.\;\ref{figure:main-structure}, for 1-bit CNNs, $\bm{x}$ is usually quantized to two numbers with the same absolute value. Thus, $p(\bm{x})$ is modeled as a Gaussian mixture with two modes:
	\begin{equation}
	\label{bczhang1}
	\begin{aligned}
	&\!p(\bm{x})\!=\!\frac{1}{2}(2\pi)^{-\frac{N}{2}}\!\det(\bm{\Psi})^{-\frac{1}{2}} \{\exp(-\frac{(\bm{x}-\bm{\mu})^T\bm{\Psi}^{-1}(\bm{x}-\bm{\mu})}{2})\\
	&+\exp(-\frac{(\bm{x}+\bm{\mu})^T\bm{\Psi} ^{-1}(\bf{x}+\bm{\mu})}{2})\}\\ 
	&\approx\!\frac{1}{2}(2\pi)^{-\frac{N}{2}}\det(\bm{\Psi})^{-\frac{1}{2}}\{\exp(-\frac{\!(\bm{x}_{+}\!-\!\bm{\mu}_{+}\!)^T\!\bm{\Psi_+}^{-1}\!(\bm{x}_{+}-\bm{\mu}_{+})}{2}\!) \\
	&+\exp(-\frac{(\bm{x}_{-}+\bm{\mu}_{-})^T\bm{\Psi_-}^{-1}(\bm{x}_{-}+\bm{\mu}_{-})}{2})\}
	\end{aligned}
	\end{equation}
	where $\bm{x}$ is divided into $\bm{x}_+$ and $\bm{x}_-$  according to the signs of the elements in $\bm{x}$ and $N$ is the dimension of $\bm{x}$. Eq. \ref{bczhang1} is obtained based on the assumption that the overlap between $\bm{x}_+$ and $\bm{x}_-$ is neglected.  Accordingly, Eq. \ref{type1} can be rewritten as:
	\begin{equation}
	\begin{aligned}
	\min &||\hat{\bm{x}}-\bm{w}\circ\bm{x}||^2_2 +\nu (\bm{x}_+-\bm{\mu}_+)^T \bm{\Psi}_+^{-1} (\bm{x}_+-\bm{\mu}_+)\\&+\nu (\bm{x}_-+\bm{\mu}_-)^T \bm{\Psi}_-^{-1} (\bm{x}_-+\bm{\mu}_-)
	+ \nu \log( \det(\bm{\Psi})),
	\end{aligned}
	\label{Bayesian kernel loss}
	\end{equation}
	where ${\bm{\mu}_-}$ and ${\bm{\mu}_+}$ are solved independently. $\det(\bm{\Psi})$ is accordingly set to be the determinant of the matrix $\bm{\Psi}_-$ or $\bm{\Psi}_+$. 
	We call Eq.\;\ref{Bayesian kernel loss} the Bayesian kernel loss.
	
	\textbf{Bayesian feature loss.} This loss is designed to alleviate the disturbance caused by the extreme quantization process in 1-bit CNNs. Considering the intra-class compactness, the features $\bm{f}_m$ of the $m$th class supposedly follow a Gaussian distribution with the mean $\bm{c}_m$  as revealed in the center loss \cite{wen2016discriminative}. Similar to the Bayesian kernel loss, we define $ \bm{y}^m_f = \bm{f}_m - \bm{c}_m$ and $ \bm{y}^m_f \sim \mathcal N(\bm{0},\bm{\sigma}_m)$, and have:
	\begin{equation}
	\begin{aligned}
	\min \!||\bm{f}_m-\bm{c}_m||^2_2\! &+\! \sum^K_{k=1}\! \left[\!\sigma_{m,k}^{-2} (f_{m,k}\! -\! c_{m,k})^2\!+\!\log(\sigma_{m,k}^2)\!\right],
	\end{aligned}
	\label{feature loss}
	\end{equation}
	which is called the Bayesian feature loss. In Eq.\;\ref{feature loss}, 
	$\sigma_{m,k}$, $f_{m,k}$ and $c_{m,k}$ are the $k$th elements of $\bm{\sigma}_{m}$, $\bm{f}_{m}$ and $\bm{c}_{m}$, respectively.
	
	\subsection{Optimized 1-bit CNNs with Bayesian learning}
	We employ the two Bayesian losses to facilitate the optimization of 1-bit CNNs. We name this method as Bayesian Optimized 1-bit CNNs (BONNs). Now, we can reformulate the two Bayesian losses for 1-bit CNNs as follows:
	\begin{equation}
	\begin{aligned}
	L_B &=\frac{\lambda}{2}\sum^{L}_{l=1}\sum^{I_l}_{i=1}\{\vert\vert \hat{\bm{X}}_i^l- \bm{w}^l \circ \bm{X}_i^l \vert\vert^2_2 \\ 
	&+\nu({\bm{X}_i^l}_+-{\bm{\mu}_i^l}_+)^T({\bm{\Psi}_i^l}_+)^{-1}({\bm{X}_i^l}_+-{\bm{\mu}_i^l}_+) \\
	&+\nu({\bm{X}_i^l}_-+{\bm{\mu}_i^l}_-)^T({\bm{\Psi}_i^l}_-)^{-1}({\bm{X}_i^l}_-+{\bm{\mu}_i^l}_-) \\
	&+ \nu\log(\det(\bm{\Psi}^l))\} +\frac{\theta}{2}\sum^M_{m=1}\{||\bm{f}_m-\bm{c}_m||^2_2 \\
	&+\sum_{k=1}^K \left[\sigma_{m,k}^{-2} (f_{m,k} - c_{m,k})^2+\log(\sigma_{m,k}^2)\right]\},  
	\end{aligned}            
	\label{bcnnlb}
	\end{equation}
	where $\bm{X}_i^l,l\in\{1,..., L\}, i\in\{1,..., I_l\}$ is the vectorization of the $i$th kernel matrix at the $l$th convolutional layer, $\bm{w^l}$ is a vector used to modulate $\bm{X}_i^l$, and $\bm{\mu}_i^l$ and $\bm{\Psi}_i^l$ are the mean and covariance of the $i$th kernel vector at the $l$th layer, respectively. Furthermore, we assume the parameters in the same kernel are independent, and thus $\bm{\Psi}_i^l$ become a diagonal matrix with the identical value $(\sigma_i^l)^2$, the variance of the $i$th kernel of the $l$th layer.
	In this case, the calculation of the inverse of $\bm{\Psi}_i^l$ is speeded up, and also all the elements of $\bm{\mu}_i^l$ are identical, equal to $\mu_i^l$. Note that in our implementation, all elements of $\bm{w}^l$ are replaced by their average during the forward process. Accordingly, only a scalar instead of a matrix is involved in the inference, and thus the computation is significantly accelerated. 
	
	In BONNs, the cross-entropy loss $L_s$, the Bayesian kernel loss and the Bayesian feature loss are aggregated together to build the total loss as:
	\begin{equation}
	L=L_S + L_{B}. 
	\label{totalloss}
	\end{equation}
	The Bayesian kernel loss constrains the distribution of the convolution kernels to a symmetric Gaussian mixture with two modes, and simultaneously, minimizes the quantization error through the $\vert\vert\hat{\bm{X}}_i^l - \bm{w}^l \circ \bm{X}_i^l\vert\vert^2_2$ term. Meanwhile the Bayesian feature loss modifies the distribution of the features to reduce the intra-class variation for better classification.
	\subsection{Backward Propagation}
	To minimize Eq.\;\ref{bcnnlb}, we update  $\bm{X}_i^l$,    $\bm{w}^l$,  $\mu_i^l$, $\sigma_i^l$, $\bm{c}_m$ and $\bm{\sigma}_m$ using the stochastic gradient descent (SGD) algorithm, which is elaborated in the following.
	
	\subsubsection{Updating $X_i^l$}
	We define $\delta_{\bm{X}_i^l}$ as the gradient of the full-precision kernel $\bm{X}_i^l$, and have:
	\begin{equation}
	\delta_{\bm{X}_i^l}=\frac{\partial L}{\partial \bm{X}_i^l}=\frac{\partial L_S}{\partial \bm{X}_i^l}+\frac{\partial L_{B}}{\partial \bm{X}_i^l}.
	\label{delta_X}
	\end{equation}
	For each term in Eq.\;\ref{delta_X}, we have:
	\begin{equation}
	\begin{aligned}
	\frac{\partial L_S}{\partial {\bm{X}}_{i}^{l}}
	&=\frac{\partial L_S}{\partial \hat{\bm{X}}_i^l} \frac{\partial \hat{\bm{X}}_i^l}{\partial (\bm{w}^l \circ \bm{X}_i^l)} \frac{\partial (\bm{w}^l \circ \bm{X}_i^l)}{\partial \bm{X}_i^l}\\
	&=\frac{\partial L_S}{\partial \hat{\bm{X}_i^l}}\circ\mathds{1}_{-1\leq\bm{w}^l\circ \bm{X}_i^l\leq1}\circ \bm{w}^l,
	\end{aligned}
	\label{Ls/X}
	\end{equation}
	
	\begin{equation}
	\begin{aligned}
	\frac{\partial L_{B}}{\partial \bm{X}_i^l} & = \lambda\{ \bm{w}^{l}\circ\left[\bm{w}^{l}\circ \bm{X}^l_i-\hat{\bm{X}}_i^l\right]\\ &	+\nu[{(\bm{\sigma}_i^l})^{-2}\circ({\bm{X}_i^l}_+-{\bm{\mu}_i^l}_+)\\ & + {(\bm{\sigma}_i^l})^{-2}\circ({\bm{X}_i^l}_-+{\bm{\mu}_i^l}_-)],
	\label{Lb/X}
	\end{aligned}
	\end{equation}
	where $\mathds{1}$ is the indicator function, which is widely used to estimate the gradient of non-differentiable parameters \cite{rastegari2016xnor}, and $(\bm{\sigma}_i^l)^{-2}$ is a vector whose all elements are equal to $(\sigma_i^l)^{-2}$.
	
	\subsubsection{Updating $\bm{w}^l$}
	Likewise,     $\delta_{\bm{w}^l}$  is composed of the following two parts:
	\begin{equation}
	\delta_{\bm{w}^l}=\frac{\partial L}{\partial \bm{w}^l}=\frac{\partial L_S}{\partial \bm{w}^l}+\frac{\partial L_B}{\partial \bm{w}^l}.
	\label{delta_w}
	\end{equation}
	For each term in Eq.\;\ref{delta_w}, we  have:
	
	\begin{equation}
	\begin{aligned}
	\frac{\partial L_S}{\partial \bm{w}^l}&=\sum^{I_l}_{i=1} \frac{\partial L_S}{\partial \hat{\bm{X}}_i^l} \frac{\partial \hat{\bm{X}}_i^l}{\partial (\bm{w}^l \circ \bm{X}_i^l)}\frac{\partial(\bm{w}^l \circ \bm{X}_i^l)}{\partial  \bm{w}^l}\\
	&=\sum^{I_l}_{i=1}\frac{\partial L_S}{\partial \hat{\bm{X}}_i^l}\circ\mathds{1}_{-1\leq\bm{w}^{l}\circ \bm{X}_i^l\leq1}\circ \bm{X}_i^l,
	\end{aligned}
	\label{Ls/w}
	\end{equation}
	\begin{equation}
	\frac{\partial L_{B}}{\partial \bm{w}^l}=\lambda\sum^{I_l}_{i=1}(\bm{w}^l \circ \bm{X}_i^l-\hat{\bm{X}}_i^l)\circ\bm{X}_i^l.
	\label{LB/w}
	\end{equation}
	
	\subsubsection{Updating $\mu_i^l$ and $\sigma_i^l$}
	Note that we use the same $\mu_i^l$ and $\sigma_i^l$ for each kernel, so the gradients here are scalars. The gradients $\delta_{\mu_i^l}$, and $\delta_{\sigma_i^l}$ are computed as:
	\begin{equation}
	\begin{aligned}
	&\delta_{\mu_i^l}=\frac{\partial L}{\partial \mu_i^l} = \frac{\partial L_{B}}{\partial \mu_i^l}\\
	&=\frac{\lambda\nu}{K_{I_l}}\sum_{k=1}^{K_{I_l}}\left\{
	\begin{aligned}
	&({\sigma_{i}^l})^{-2}({\mu_{i}^l}-{X_{i,k}^l}), & X_{i,k}^l \ge 0, \\
	&({\sigma_{i}^l})^{-2}({\mu_{i}^l}+{X_{i,k}^l}), & X_{i,k}^l < 0, \\
	\end{aligned}
	\right.
	\end{aligned}
	\label{delta_mu}
	\end{equation}
	
	\begin{equation}
	\begin{aligned}
	\delta_{\sigma_i^l} &= \frac{\partial L}{\partial \sigma_i^l} = \frac{\partial L_{B}}{\partial \sigma_i^l} \\
	&=\frac{\lambda\nu}{K_{I_l}}\!\sum^{K_{I_l}}_{k=1}\!\left\{\!
	\begin{aligned}
	&\!-\!({\sigma_{i}^l})^{-3}({X_{i,k}^l}-{\mu_{i}^l})^2\!+\! ({\sigma_{i}^l})^{-1}, X_{i,k}^l\!\ge\! 0, \\
	&\!-\!({\sigma_{i}^l})^{-3}({X_{i,k}^l}+{\mu_{i}^l})^2\!+\! ({\sigma_{i}^l})^{-1}, X_{i,k}^l\!<\! 0, \\
	\end{aligned}
	\right.	
	\end{aligned}
	\label{delta_Psi}
	\end{equation}
	where $X_{i,k}^l,k\in\{1,...,K_{I_l}\}$, denotes the $k$th element of vector $\bm{X}_i^l$.
	We update $\bm{c}_m$ using the same strategy as the center loss \cite{wen2016discriminative} in the fine-tuning process, while updating $\sigma_{m,k}$ based on $L_B$ is straightforward, which is not elaborated here for brevity. The above equations show that the proposed method is trainable in an end-to-end manner. Finally we summarize the whole learning procedure in Algorithm \ref{alg:algorithm}.
	
	\begin{algorithm}[tb] 
		\small
		\caption{Optimized 1-bit CNN with Bayesian learning} 
		\label{alg:algorithm} 
		\begin{algorithmic}[1] 
			\REQUIRE ~~\\ 
			The training dataset; the full-precision kernels $\bm{X}$; the modulation vector $\bm{w}$; the learning rate $\eta$, regularization parameter $\lambda, \theta$ and variance $\nu$.
			\ENSURE ~~\\ 
			The BONN based on the updated $\bm{X}$, $\bm{w}$, $\bm{\mu}$, $\bm{\sigma}$, $\bm{c}_m$, $\bm{\sigma}_m$.
			\STATE Initialize $\bm{X}$ and $\bm{w}$ randomly, and then estimate $\bm{\mu}$, $\bm{\sigma}$ based on the average and variance of $\bm{X}$, respectively; 
			\REPEAT
			\STATE // Forward propagation
			\FOR {$l=1$ to $L$}
			\STATE $\hat{\bm{X}}_{i}^l = \bm{w}^l \circ {\rm sign}({\bm{X}}_{i}^l), \forall i$; //  Each element of $\bm{w}^l$ is replaced by the average of all elements.
			\STATE Perform activation binarization;	// Using the sign function
			\STATE Perform 2D convolution with $\hat{\bm{X}}_i^l$, $ \forall i$;
			\ENDFOR
			\STATE // Backward propagation
			\STATE Compute $\delta_{\hat{\bm{X}}_{i}^l}=\frac{\partial L_s}{\partial \hat{\bm{X}}_{i}^l}, \forall  l,i$;
			\FOR {$l=L$ to $1$} 
			\STATE Calculate $\delta_{\bm{X}_i^l}$, $\delta_{\bm{w}^l}$, $\delta_{\mu_i^l}$, $\delta_{\sigma_i^l}$; // using Eqs.\;\ref{delta_X}$\sim$\ref{delta_Psi}
			\STATE  Update parameters $\bm{X}_i^l, \bm{w}^l, \mu_i^l, \sigma_i^l$ using SGD;
			\ENDFOR
			\STATE Update $\bm{c}_m, \bm{\sigma}_m$;
			\UNTIL {the algotirhm converges.}
		\end{algorithmic}
	\end{algorithm}
	
	\section{Experiments}
	We perform the image classification task on the CIFAR-10/100 \cite{krizhevsky2014cifar} and ILSVRC12 ImageNet datasets \cite{deng2009imagenet} to evaluate the performance of BONNs. Considering the favorable generalization capability of our method, BONNs could be integrated in any DCNN variants. For a fair comparison with other state-of-the-art 1-bit CNNs, we apply Wide-Resnet (WRN) \cite{zagoruyko2016wide} and ResNet18 \cite{he2016deep} as the full-precision backbone networks. In the following experiments, both the kernels and the activations are binarized. The leading performances reported in the following sections verify the superiority of our BONNs.
	
	\subsection{Datasets and Implementation Details}
	\subsubsection{Datasets}
	CIFAR-10 \cite{krizhevsky2014cifar} is a natural image classification dataset, composed of a training set and a test set, each with 50,000 and 10,000 32$\times$32 color images, respectively. These images span across 10 different classes, including airplanes, automobiles, birds, cats, deer, dogs, frogs, horses, ships and trucks. 
	Comparatively, CIFAR-100 is a more comprehensive dataset containing 100 classes. On CIFAR-10/100, WRNs are employed as the backbones of BONNs. In comparison, ILSVRC12 ImageNet object classification dataset \cite{deng2009imagenet} is more diverse and challenging. It contains 1.2 million training images, and 50,000 validation images, across 1000 classes. For comparing with other state-of-the-art methods on this dataset, we adopt ResNet18 as the backbone to verify the effectiveness and superiority of our BONNs.
	
	\subsubsection{WRN}
	The structure of WRN is similar to ResNet in general. Yet additionally, a depth factor $k$ is introduced to control the feature map depth expansion through 3 stages, while the spatial dimension of the features is kept the same. For brevity, we set $k$ to 1 in the experiments. Besides, the number of channels in the first stage is another important parameter in WRN. We set it to 16 and 64, thus resulting in two network configurations: 16-16-32-64 and 64-64-128-256. In the 64-64-128-256 network, a dropout layer with a ratio of 0.3 is added to prevent overfitting. The learning rate is initially set to 0.01,  which decays by 20\% per 60 epochs until reaching the maximum epoch of 200 on CIFAR-10/100. We set $\nu$ to $1e-4$ for quantization error in WRN. Bayesian feature loss is only used in the fine-tuning process. Other training details are the same as those described in \cite{zagoruyko2016wide}. WRN-22 denotes a network with 22 convolutional layers and similarly for WRN-40. 
	
	\subsubsection{ResNet18}
	
	For ResNet18, we binarize the features and kernels in the backbone convolution layers without convolution layers in shortcuts, following the settings and network modifications in Bi-Real Net \cite{liu2018bi}. The SGD algorithm is with a momentum of 0.9 and a weight decay of $1e-4$. The learning rate for $\bm{w}^l, \sigma_i^l$ is set to 0.01, while for $\bm{X}^l_i, \mu_i^l$ and other parameters the rates are set to 0.1. $\nu$ is set to $1e-3$ for the quantization error in ResNet18. The strategy of the learning rate decay is also employed, which is a degradation of 10\% for every 30 epochs before the algorithm reaches the maximum epoch of 70.  	
%
	
	\begin{figure}[tb]
		\centering
		\includegraphics[width=\linewidth]{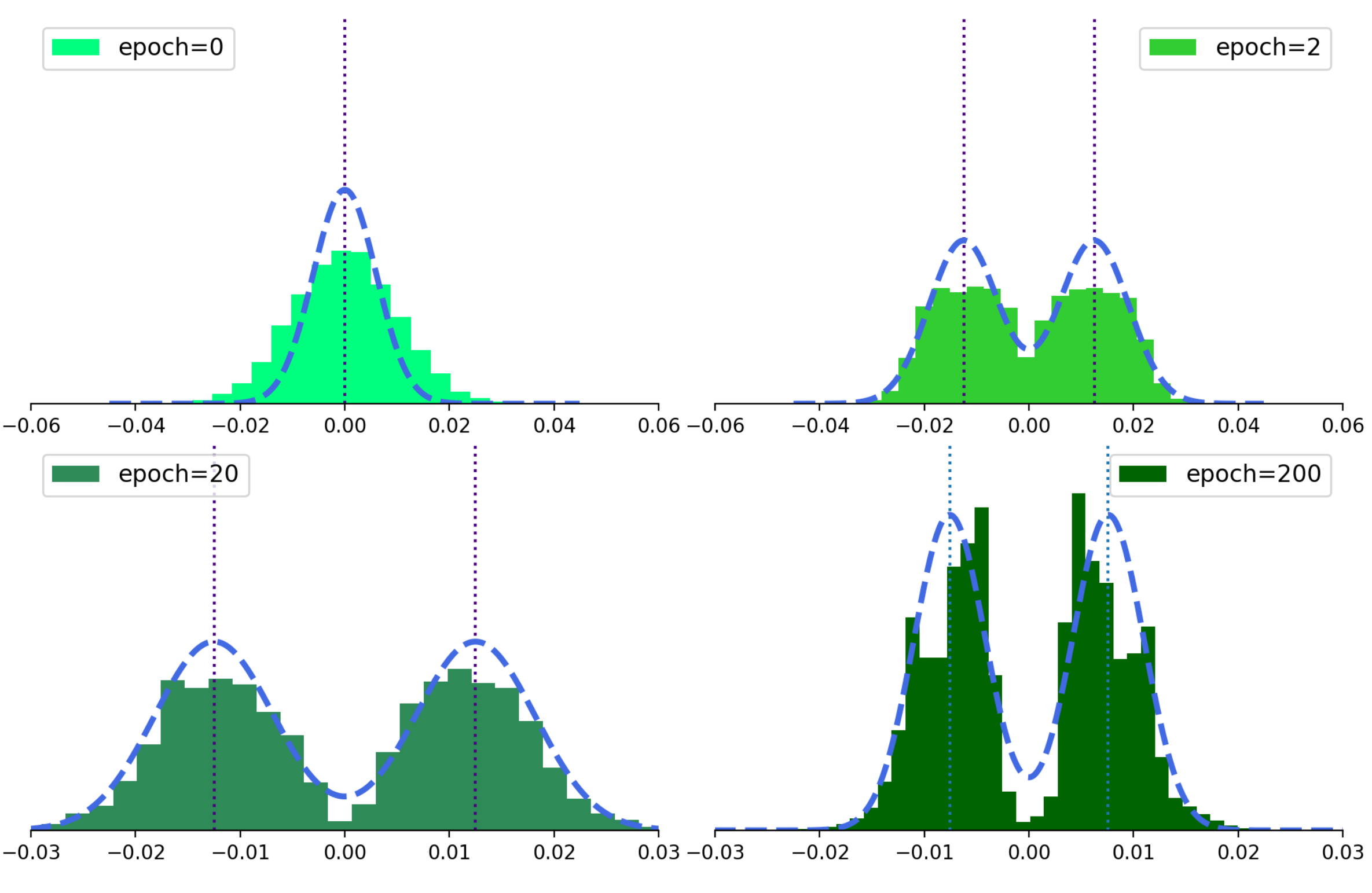}
		\caption{We demonstrate the kernel weight distribution of the first binarized convolution layer of BONNs. Before training, we initialize the kernels as a single-mode Gaussian distribution. From the $2$th epoch to the $200$th epoch, with $\lambda$ fixed to $1e-4$, the distribution of the kernel weights becomes more and more compact with two modes, which confirms that the Bayesian kernel loss can regularize the kernels into a promising distribution for binarization.}  
		\label{figure:lambda}
	\end{figure}
	
	\begin{figure}[tb]
		\centering
		\includegraphics[width=\linewidth]{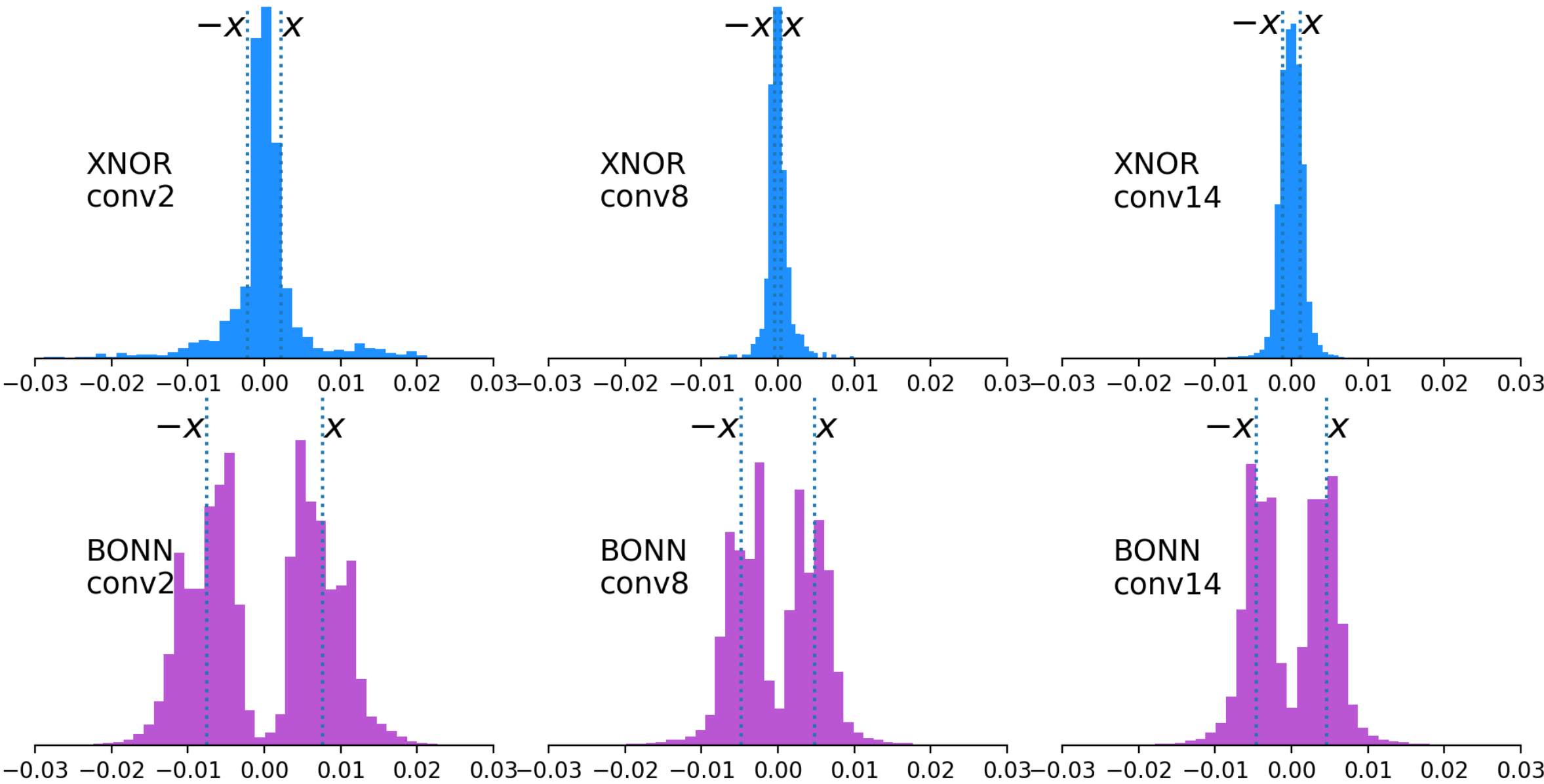}
		\caption{Weight distribution of XNOR and BONN, both based on WRN22 (2nd, 8th and 14th convolutional layers) after 200 epochs. The weight distribution difference between XNOR and BONN indicates that the kernels are regularized with our proposed Bayesian kernel loss, across the convolutional layers.}  
		\label{figure:distribution}
	\end{figure}
	
	\begin{figure}[tb]
		\centering
		\includegraphics[width=\linewidth]{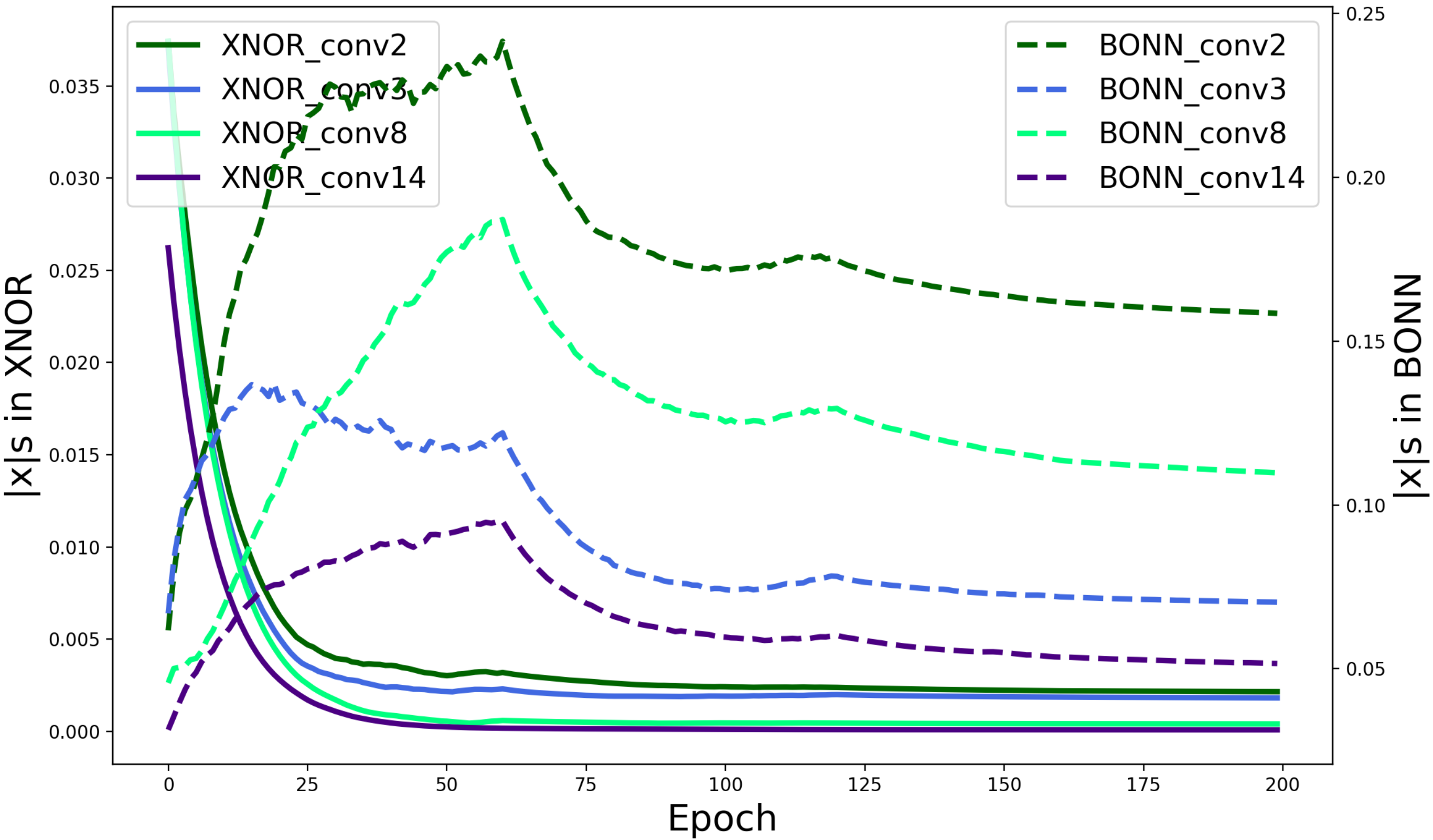}
		\caption{The evolution of the binarized values, $|x|$s, during the training process of XNOR and BONN. They are both based on WRN22 (2nd, 3rd, 8th and 14th convolutional layers) and the curves are not sharing the same y-axis. The binarized values of XNOR Net tend to converge to small and similar values but these of BONN are learned diversely.}  
		\label{figure:epoch}
	\end{figure}

	\begin{table}[tb]
		\caption{Effect of using or not using the Bayesian losses on the ImageNet dataset. The backbone is ResNet18.}
		\centering
		\begin{tabular}{c c c c c c }
			\toprule
			\multicolumn{2}{c}{Bayesian kernel loss} &   & \checkmark &  & \checkmark   \\
			\cmidrule(lr){3-6}
			\multicolumn{2}{c}{Bayesian feature loss} &  &  & \checkmark & \checkmark  \\
			\hline
			\multirow{2}{*}{Accuracy}& Top-1 & 56.3 & 58.3 & 58.4 & \textbf{59.3} \\
			\cmidrule(lr){2-6}
			& Top-5 & 79.8 & 80.8 & 80.8 &\textbf{81.6} \\
			\bottomrule
		\end{tabular}
		\label{table:ablation}
	\end{table}
	
	\begin{table*}[tb]
		\caption{Test accuracies on the CIFAR-10/100 datasets. BONNs are based on WRNs \cite{rastegari2016xnor}. We calculate the number of parameters for each model and the numbers refer to the models on CIFAR-10. Note that for the full-precision models, each parameter takes 32 bits, while for the binary models, each takes only 1 bit.}
		\centering
		\small
		\begin{tabular}{c c c c c c}
			\toprule
			\multirow{2}{*}{\large Model} & \multirow{2}{*}{\large Kernel stage} & \multirow{2}{*}{\large \#Param} & \multicolumn{2}{c}{Dataset} \\
			\cmidrule(lr){4-5}
			&&& CIFAR-10 & CIFAR-100 \\
			\hline
			\hline
			WRN22 & 16-16-32-64 & 0.27M & 91.66 & 67.51 \\
			XNOR-Net & 16-16-32-64  & 0.27M & 81.90 & 53.17 \\
			BONN & 16-16-32-64 & 0.27M & \textbf{87.34} & \textbf{60.91} \\
			\hline
			WRN22 & 64-64-128-256 & 4.33M & 94.96 & - \\
			XNOR-Net & 64-64-128-256 & 4.33M & 88.52 & - \\
			BONN & 64-64-128-256 & 4.33M & \textbf{92.36} & - \\
			\bottomrule
		\end{tabular}
		\label{table:cifar}
	\end{table*}
	\subsection{Ablation Study}
	In this section, we evaluate the effects of the hyper-parameters on the performance of BONNs, including $\lambda$ and $\theta$. The Bayesian kernel loss and the Bayesian feature loss are balanced by $\lambda$ and $\theta$, respectively, for adjusting the distributions of kernels and features in a better form. CIFAR-10 and WRN22 are used in the experiments. The implementation details are given below.
	
	We first vary $\lambda$ and also set it to zero for validating the influence of the Bayesian kernel loss on the kernel distribution. The utilization of the Bayesian kernel loss effectively improves the accuracy on CIFAR-10. But the accuracy does not increase with $\lambda$, which indicates what we need is not a larger $\lambda$, but a proper $\lambda$ to  reasonably balance the relationship between the cross-entropy loss and the Bayesian kernel loss. For example, when $\lambda$ is set to $1e-4$, we obtain an optimal balance and the classification accuracy is the best.
	
	The hyper-parameter $\theta$ dominates the intra-class variations of the features, and the effect of the Bayesian feature loss on the features is also investigated by changing $\theta$. The results illustrate that the classification accuracy varies in a way similar to $\lambda$, which verifies that the Bayesian feature loss can lead to a better classification accuracy when a proper $\theta$ is chosen.
	
	To understand the Bayesian losses better, we carry out an experiment to examine how each loss affects the performance. According to the above experiments, we set $\lambda$ to $1e-4$ and $\theta$ to $1e-3$, if they are used. As shown in Table \ref{table:ablation}, both the Bayesian kernel loss and the Bayesian feature loss can independently improve the accuracy on ImageNet, and when applied together, the Top-1 accuracy reaches the highest value of 59.3\%.
	
	Besides, Fig.\;\ref{figure:lambda} illustrates the distribution of the kernel weights, with $\lambda$ fixed to $1e-4$. During the training process, the distribution is gradually approaching the two-mode GMM as assumed previously, confirming the effectiveness of the Bayesian kernel loss in a more intuitive way. We also compare the kernel weight distribution between XNOR Net and BONN. As is shown in Fig. \ref{figure:distribution}, the kernel weights learned in XNOR Net distribute tightly around the threshold value but these in BONN are regularized in a two-mode GMM style. The Fig. \ref{figure:epoch} shows the  evolution of the binarized values during the training process of XNOR Net and BONN. The two different pattern indicates the binarized values learned in BONN are more diverse.
	
	\subsection{Results on the CIFAR-10/100 datasets}
	We first evaluate our proposed BONNs in comparison with XNOR-Net \cite{rastegari2016xnor} with WRN backbones and also report the accuracy of full-precision WRNs on CIFAR-10 and CIFAR-100. Three WRN variants are chosen for a comprehensive comparison: 22-layer WRNs with the kernel stage of 16-16-32-64 and 64-64-128-256. We also use data augmentation where each image is with a padding size of $4$ and is randomly cropped into $32\times32$ for CIFAR-10/100. Table \ref{table:cifar} indicates that BONNs outperform XNOR-Net on both datasets by a large margin in all the three cases. Compared with the full-precision WRNs, BONNs eliminate the accuracy degradation to an acceptable level, {\it e.g.}, only 2.6\% left on the backbone WRN22 with 64-64-128-256, which verifies the advantage of our method in building 1-bit CNNs. 
	
	
	\begin{figure}[tb]
		\centering
		\includegraphics[width=\linewidth]{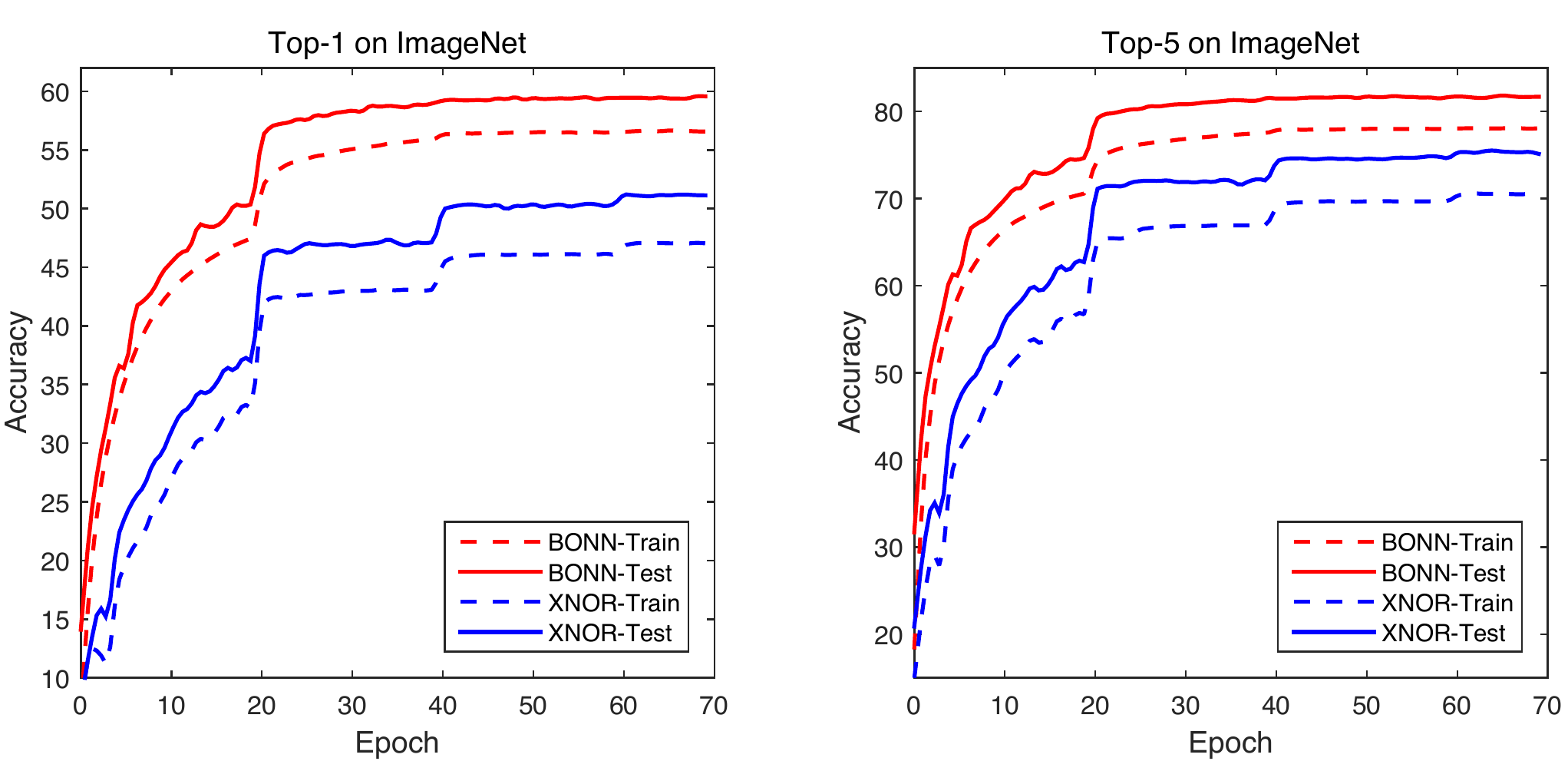}
		\caption{Training and Test accuracies on ImageNet when $\lambda=1e-4$, which shows the superiority of the proposed BONN over XNOR-Net. The backbone of the two networks is ResNet18.}
		\label{figure:acccurve}
	\end{figure}
	\begin{table}[tbp]
		\caption{Test accuracies on ImageNet. 'W' and 'A' refer to the weight and activation bitwidth respectively. The backbone of all the models is ResNet18.}
		\centering
		\begin{tabular}{c c c c c}
			\toprule
			Model & W & A & Top-1 & Top-5 \\
			\hline
			\hline
			ResNet18 & 32 & 32 & 69.3 & 89.2 \\
			BWN & 1 & 32 & 60.8 & 83.0 \\
			DoReFa-Net & 1 & 4 & 59.2 & 81.5 \\
			TBN & 1 & 2 & 55.6 & 79.0\\
			BNN & 1 & 1 & 42.2 & 67.1 \\
			XNOR-Net & 1 & 1 & 51.2 & 73.2 \\
			ABC-Net & 1 & 1 & 42.7 & 67.6 \\
			Bi-Real Net & 1 & 1 & 56.4 & 79.5 \\
			PCNN & 1 & 1 & 57.3 & 80.0 \\
			\hline
			BONN & 1 & 1 & \textbf{59.3} & \textbf{81.6} \\
			\bottomrule
		\end{tabular}
		\label{table:imagenet}
	\end{table}
	\subsection{Results on the ImageNet dataset}
	To further evaluate the performance of our method, we evaluate BONNs on the ImageNet dataset. Fig.\;\ref{figure:acccurve} shows the curves of the Top-1 and Top-5 training/test accuracies. Notably, we adopt two data augmentation methods in the training set: 1) cropping the image to the size of 224$\times$224 at a random location, and 2) flipping the image horizontally. In the test set, we simply crop the image to 224$\times$224 in the center. ResNet18 is the backbone, only with slight structure adjustments as described in \cite{liu2018bi}. 
	
	We compare the performance of BONN with other state-of-the-art quantized networks, including BWN \cite{rastegari2016xnor}, DoReFa-Net \cite{zhou2016dorefa}, TBN \cite{Wan_2018_ECCV}, XNOR-Net \cite{rastegari2016xnor}, ABC-Net \cite{lin2017towards}, BNN \cite{paper10}, Bi-Real Net \cite{liu2018bi} and PCNN \cite{gu2018projection}. Table \ref{table:imagenet} indicates that our BONN obtains the highest accuracy among these 1-bit CNNs, in which Bi-Real Net and PCNN perform most similar to BONN, yet BONN outperforms them by about 3\% and 2\% in Top-1 accuracy, respectively. Moreover, due to the application of the clip function \cite{liu2018bi}, Bi-Real Net is trained in a two-stage procedure which requires an extra cost. It is also worth mentioning that DoReFa-Net and TBN use more than 1-bit to quantize the activations, yet we still get better performance in comparison. These results show that BONNs are not limited to small datasets, but also work well on large datasets. This further verifies the generalization capability of our BONNs.
	
	\subsection{Memory Usage and Efficiency Analysis}
	In a full-precision network, each parameter requires 32 bits to save it. While in 1-bit CNNs, each parameter is stored with just 1 bit. In BONNs, we follow a strategy adopted by XNOR-Net, which keeps full-precision the parameters in the first convolution layer, all $1\times1$ convolutions and the fully connected layer. This leads to an overall compression rate of 11.10 for ResNet18. For efficiency analysis, if all of the operands of the convolutions are binary, then the convolutions can be estimated by XNOR and bit-counting operations.\cite{paper10}. In this way, we can get $58\times$ faster on CPUs \cite{rastegari2016xnor}. 
	\section{Conclusion and future work}
	In this paper, we have proposed Bayesian optimized 1-bit CNNs (BONNs), which take the full-precision kernel and feature distributions into consideration, resulting in a unified Bayesian framework with two new Bayesian losses. The Bayesian losses are used to adjust the distributions of kernels and features towards an optimal solution. Comprehensive studies on the hyper-parameters for the Bayesian losses have been conducted. Extensive experiments on CIFAR and ImageNet demonstrate that BONNs achieve the best classification performance for WRNs and ResNet18, and have superior performance over other 1-bit CNNs. 
	In the future, we plan to explore our proposed BONNs on deeper networks like ResNet34 and on different tasks other than classification.
	\section{Acknowledgement}
	
	The work is supported by the Fundamental Research Funds for Central Universities,  National Natural Science Foundation of China under Grant  61672079, in part by Supported by Shenzhen Science and Technology Program (No.KQTD2016112515134654).
	{\small
 		\bibliographystyle{ieee_fullname}
		\bibliography{bibliography}
	}
	
\end{document}